\documentclass{article}

\PassOptionsToPackage{numbers, sort, compress}{natbib}

    \usepackage[preprint]{neurips_2025}

\usepackage[utf8]{inputenc} 
\usepackage[T1]{fontenc}    
\usepackage{hyperref}       
\usepackage{url}            
\usepackage{booktabs}       
\usepackage{amsfonts}       
\usepackage{nicefrac}       
\usepackage{microtype}      
\usepackage{xcolor}         
\usepackage{graphicx}

\usepackage{multirow}
\usepackage{amsmath} 
\usepackage{xcolor}  
\usepackage{pifont}
\usepackage{enumitem}
\newcommand{\with}{\textcolor{black}{\ding{51}}}

\definecolor{MyDarkGreen}{RGB}{0, 100, 0} 

\title{3DFlowAction: Learning Cross-Embodiment Manipulation from 3D Flow World Model}

\author{
    Hongyan Zhi\textsuperscript{\rm 1}\thanks{Equal contribution.}~~\thanks{Work done during the internship at Tencent Robotics X} ~~
    Peihao Chen\textsuperscript{\rm 2}\footnotemark[1] ~~ 
    Siyuan Zhou\textsuperscript{\rm 3}\footnotemark[1] ~~ 
    Yubo Dong\textsuperscript{\rm 1} ~~ \\
    \textbf{Quanxi Wu}\textsuperscript{\rm 1} ~~ 
    \textbf{Lei Han}\textsuperscript{\rm 2} ~~
    \textbf{Mingkui Tan}\textsuperscript{\rm 1 \rm 4} \thanks{Corresponding author.} ~~ \\
    \textsuperscript{\scriptsize{\rm 1}}\small{South China University of Technology}
    \textsuperscript{\scriptsize{\rm 2}}\small{Tencent Robotics X} \\
    \textsuperscript{\scriptsize{\rm 3}}\small{Hong Kong University of Science and Technology}
    \textsuperscript{\scriptsize{\rm 4}}\small{Pazhou Laboratory}
}

\begin{document}

\maketitle

\begin{abstract}
Manipulation has long been a challenging task for robots, while humans can effortlessly perform complex interactions with objects, such as hanging a cup on the mug rack. A key reason is the lack of a large and uniform dataset for teaching robots manipulation skills. Current robot datasets often record robot action in different action spaces (\textit{e.g.} joint angles or end-effector pose in different base coordinates) within a simple scene. This hinders the robot to learn a unified and robust action representation for different robots within diverse scenes.
Observing how humans understand a manipulation task, we find that understanding how the objects should move in the 3D space is a critical clue for guiding actions. This clue is embodiment-agnostic and suitable for both humans and different robots.
Motivated by this, we aim to learn a 3D flow world model from both human and robot manipulation data. This model predicts the future movement of the interacting objects in 3D space, guiding action planning for manipulation.
Specifically, we synthesize a large-scale 3D optical flow dataset, named ManiFlow-110k, through a moving object auto-detect pipeline. A video diffusion-based world model then learns manipulation physics from these data, generating 3D optical flow trajectories conditioned on language instructions.
With the generated 3D object optical flow, we propose a flow-guided rendering mechanism, which renders the predicted final state and leverages GPT-4o to assess whether the predicted flow aligns with the task description. This equips the robot with a closed-loop planning ability.
Finally, we consider the predicted 3D optical flow as constraints for an optimization policy to determine a chunk of robot actions for manipulation.
Extensive experiments demonstrate strong generalization across diverse robotic manipulation tasks and reliable cross-embodiment adaptation without hardware-specific training. Code and data will be available at  \url{ https://github.com/Hoyyyaard/3DFlowAction/ }

\end{abstract}

\begin{figure}[th]
  \centering
\includegraphics[width=1\linewidth]{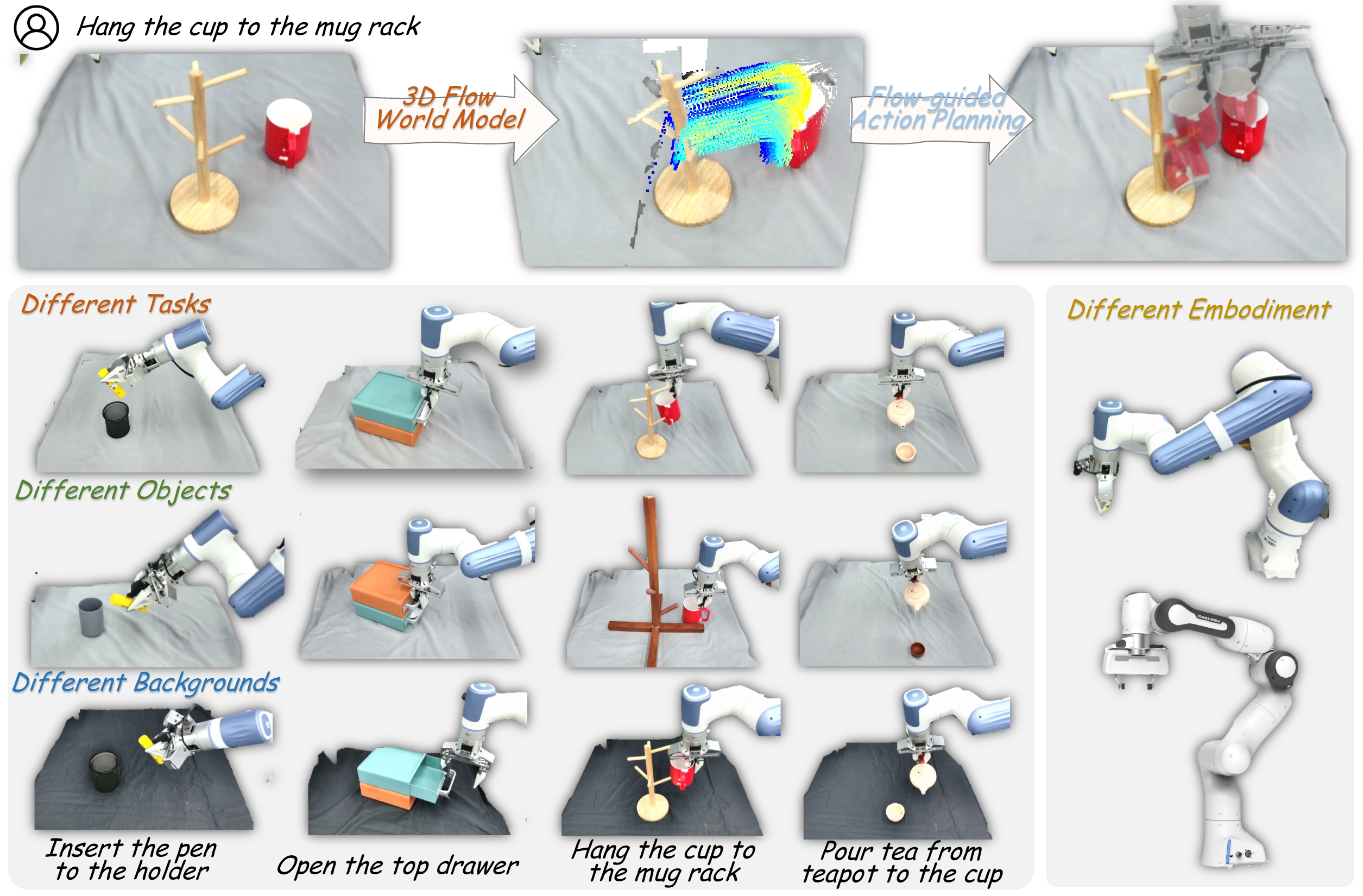}
  \caption{3DFlowAction seeks to build a flow world model to generate 3D optical flow that serves as action guidance for downstream manipulation tasks. Experiments on four complex foundational tasks in different settings demonstrate strong generalization across various manipulation tasks and reliable cross-embodiment adaptation without hardware-specific training.}
  \label{fig::teaser}
\end{figure}

\section{Introduction}

Manipulation is a fundamental capability for intelligent agents~\cite{fang2023anygrasp, black2024vision}. While humans demonstrate remarkable dexterity in object interaction, robotic systems remain a persistent challenge in unstructured environments. A central challenge lies in the scarcity of large-scale, unified training datasets that capture the physical and semantic complexity required for generalizable manipulation learning, hindering progress in robotic skill acquisition. However, collecting operational data from robotic arms requires significant human and material resources, making it difficult to scale. Therefore, learning manipulation skills from human videos has emerged as a promising approach to advance this field.
This raises an important question: \textit{How can we derive a unified and embodiment-agnostic action representation from video data?}

By studying how humans perceive a manipulation task, we imagine future changes in the scene based on the task information before executing the action. The world model, which learns the dynamics of physical motion from large-scale video data and simulates future states based on instructions~\cite{bar2024navigation}, is a reliable approach to guiding robotic arm movements.
Existing methods~\cite{ko2023learning, bar2024navigation, bharadhwaj2024gen2act} focus on learning a video world model from video data to guide robotic movements. However, this framework is not object-centric, as it needs to account for irrelevant content such as background and robot arm, which leads to poor generalization. Furthermore, the video world model's planning for future states is primarily confined to a 2D plane, making it challenging to accurately represent objects' movement in 3D space.


Motivated by this, we aim to learn a 3D flow world model from human and robot manipulation videos, which predicts the future trajectory of the manipulated object as a motion cue for action policy.  Then, a flow-guided rendering machine is proposed to render the final scene state, using GPT-4o~\cite{hurst2024gpt} to assess the accuracy of the optical flow predictions, thereby achieving closed-loop planning. Finally, we treat the predicted 3D optical flow as constraints for an optimization policy that determines a sequence of robot actions for downstream manipulation, without the need for any robot action labels.

Existing human and robot manipulation video datasets often feature cluttered backgrounds and identical objects, complicating the use of standard detectors for obtaining masks of manipulated objects. To address this, we propose a moving object detection pipeline that accurately identifies these objects, ensuring valid optical flow labels and resulting in the creation of the clean 3D flow demonstration dataset, ManiFlow-110k.
We then utilize a video diffusion model to build our flow world model, which learns object motion patterns based on large-scale optical flow demonstrations, designed to be object-centric and embodiment-agnostic. Additionally, we introduce a flow-guided rendering machine to enhance optical flow prediction reliability by estimating a transformation matrix from selected optical flow points in the first and last frames. This allows us to render predicted final positions of 3D object points, which we input into GPT-4o to verify alignment with given instructions, enabling closed-loop planning. 
Furthermore, we generate a task-relevant grasp pose by utilizing the transformation matrix to avoid unreachable target positions for the robot arm.
Finally, we use the predicted 3D flow as a constraint for an optimization policy, minimizing the positions of optical flow points corresponding to the same object across time steps to derive final actions without requiring annotated supervision data.
In summary, our contributions are threefold.
\begin{itemize}[leftmargin=*]
\item We propose 3DFlowAction that utilizes 3D optical flow as a robust and unified action representation for robot planning. By learning a 3D flow world model from video data, we can simulate future object trajectories, providing object-centric and embodiment-agnostic motion cues essential for action policies. 
\item We introduce a flow-guided action policy that utilizes a flow condition optimization procedure for cross-embodiment action generation. Since 3D flow accurately represents the motion of objects, we simulate the future state of the object and employ a VLM to assess the accuracy of the predicted 3D flow for closed-loop planning. We also leverage this characteristic and the inverse kinematics of the robot arm to generate a reasonable grasp pose.

\item Extensive experiments demonstrate our method’s robust generalization across a wide range of complex robotic manipulation tasks, as well as reliable cross-embodiment adaptation without the need for hardware-specific training.
\end{itemize}

\section{Related works}
\noindent\textbf{Imitation Learning for Manipulation}. Imitation learning offers an effective approach for robots to develop human-like skills, usually depending on a large number of expert demonstrations. While 2D image-based policies that directly map the visual input to actions~\cite{florence2022implicit, chi2023diffusion, pari2021surprising, haldar2023teach} have predominated the field. With the rapid development of Vision-Language-Models~\cite{beyer2024paligemma, karamcheti2024prismatic}, Vision-Language-Action-Models(VLAs)~\cite{kim2024openvla, black2024vision} explore how to combine visual perception, language understanding, and action generation into a unified model. Despite their emergence, VLAs demand extensive robotic datasets and still encounter challenges with generalization.

\noindent\textbf{Video world model for Manipulation.}
Recent advancements in video generation models have positioned them as promising world models for robotics~\cite{wu2024ivideogpt,yang2023learning,zhou2024robodreamer,chi2024eva}.
UniPi~\cite{du2023learning} employs a video diffusion model to predict future frames and infers actions by the inverse dynamic policy, while AVDC~\cite{ko2023learning} utilizes pre-trained video prediction networks to predict the video to estimate actions directly.
Uni-Sim~\citep{yang2023learning} scales up the world models in the large dataset and demonstrates the utility of world models for policy learning. TesserAct~\cite{zhen2025tesseract} simulates 3D dynamics by predicting RGB, depth, and normal videos. However, video models remain constrained by prohibitive computational demands, limiting their practical deployment in real-world robotics applications.


\noindent\textbf{Optical flow for manipulation.}
Optical flows represent the future paths of query points on images or point clouds, which serve as universal descriptors for motion in videos. Early works~\cite{yuan2024general, wen2023any} first inject the optical flow into the robotics world. Some studies~\cite{bharadhwaj2024track2act, tang2024embodiment} propose learning image flow prediction from goal image guidance. Recently, Gen2Act~\cite{bharadhwaj2024gen2act} and FLIP~\cite{gaoflip} explore the flow-guided video generation. Im2Flow2Act~\cite{xu2024flow} and General-Flow~\cite{yuan2024general} directly learn a flow generator from human-collected collected and guides the action planning. However, they mostly focus on the application of 2D optical flow, which cannot fully describe the motion of objects in 3D space. Additionally, the limited diversity of training data results in poor generalization.

\begin{figure}[t]
  \centering
  \includegraphics[width=0.95\linewidth]{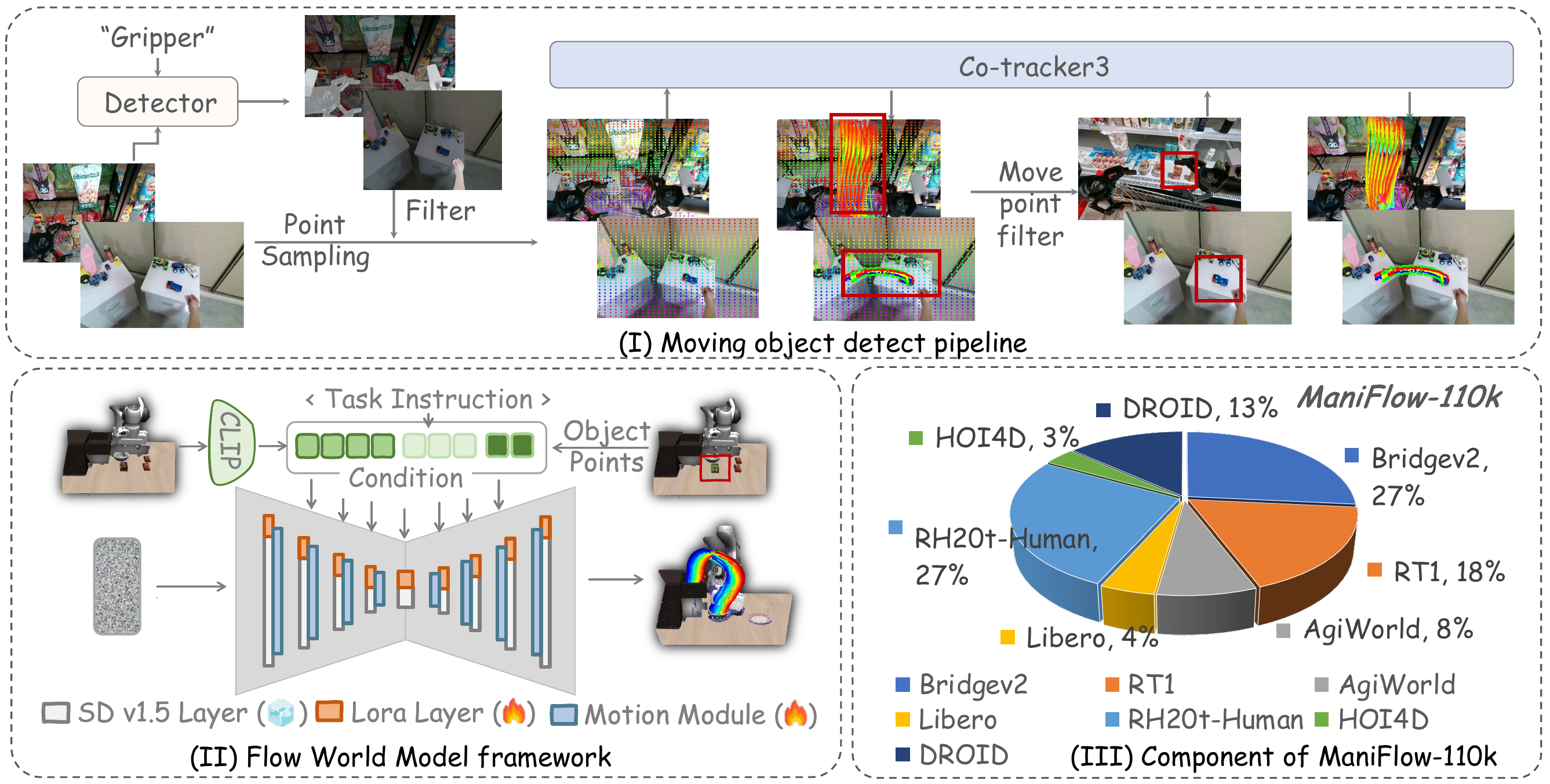}
  \caption{\textbf{Overview of 3D Flow Generation pipeline}. (I) We synthesized the 3D flow dataset ManiFlow-110k using a moving object detection pipeline. (II) We pre-trained a video diffusion model as the flow world model on ManiFlow-110k to learn the physical motion patterns of objects in manipulation tasks. (III) ManiFlow-110k comes from a wide range of robot and human videos.}
  \label{fig::img2flow}
\end{figure}

\section{3D Flow World Model for Manipulation} \label{sec::method}

Predicting optical flow as a representation of object movement has been explored in previous work~\cite{xu2024flow, wen2023any, bharadhwaj2024track2act}. However, most of these studies have primarily focused on generating 2D optical flow and have been trained on downstream manipulation datasets. 2D optical flow is inadequate for fully capturing an object's motion in 3D space, including rotation and movement perpendicular to the camera. Furthermore, training on limited datasets often leads to challenges in generalizing across diverse scenes. Therefore, we are dedicated to developing a 3D flow world model that learns to simulate object motions based on various instructions through large-scale pretraining.

\subsection{Extracting 3D Flow from Raw Video}
The 3D flow world model requires learning motion patterns from a substantial amount of data involving object movement. We have decided to construct our dataset from multiple open-source robotic and human videos. However, these open-source datasets often contain cluttered backgrounds and may include other objects that resemble the manipulated objects, making it challenging to identify the target objects accurately. Existing detection models~\cite{ren2024grounding, ravi2024sam, cheng2024yolo} have shown suboptimal performance in this context. To address this issue, we propose a moving object detection pipeline that can automatically identify manipulated objects in video data while avoiding the extraction of optical flow from the gripper. We validated the accuracy of this pipeline on the BridgeV2~\cite{walke2023bridgedata} dataset, and the results indicate that our proposed pipeline achieves a moving object detection accuracy of over 80\%.

Specifically, as shown in Fig.~\ref{fig::img2flow}(I), we first utilize Grounding-SAM2~\cite{ren2024grounding} to segment the robotic gripper mask from the first frame of the video. We then generate a set of points distributed across the entire first frame and exclude points that fall within the gripper mask. After obtaining the initial points, we employ the 2D tracking model Co-tracker3~\cite{karaev2024cotracker3} to track the motion of these points, ultimately identifying those that exhibit significant movement within the video. We select the maximum bounding box around these points to define the location of the moving objects. Once the moving objects are identified, we again use Co-tracker3 to extract the 2D optical flow of the objects' motion. We also remove the camera motion follow~\cite{yang2025magma} if needed. Following the acquisition of the 2D optical flow, we apply DepthAnythingV2~\cite{yang2024depth} to perform depth prediction on the video and project the 2D optical flow into 3D, resulting in the final 3D optical flow we require. We ultimately generated 110k instances of 3D optical flow data from open source datasets~\cite{walke2023bridgedata, yu2023scaling, khazatsky2024droid, fang2023rh20t, liu2023libero, bu2025agibot} for pre-training, named \textbf{ManiFlow-110k}, as shown in Fig.~\ref{fig::img2flow}(III).


\subsection{Learning 3D Flow World Model}
Our ultimate goal is to generate the motion trajectory of the target object, conditioning on the operational task instructions and the current scene state in 3D space. We follow the approach outlined in Im2Flow2Act~\cite{xu2024flow} and utilize AnimateDiff~\cite{guo2023animatediff} as our optical flow generator.

The proposed flow generator, denoted as $G$, generates time-varying 3D flow $\mathcal{F}$ conditioned on the initial RGB observation, task prompt, and initial points $\mathcal{F}_0$, as shown in Fig.~\ref{fig::img2flow}(II).
To achieve this, we utilize the CLIP encoder~\cite{radford2021learning} to encode the RGB observation and the task prompt while employing sinusoidal positional encoding for initial points $\mathcal{F}_0$.
Unlike Im2Flow2Act, we do not compress the 3D flow into latent space, as we discovered that the image VAE from StableDiffusion~\cite{rombach2022high} struggles to encode depth information effectively, even after fine-tuning. Consequently, we bypass the VAE and directly input the 3D flow into the U-Net.
Specifically, we define our 3D optical flow as $\mathcal{F} \in \mathbb{R}^{T \times H \times W \times 4}$, where the first two channels represent 2D coordinates in image space, the third channel denotes depth, and the fourth denotes visibility.
We follow AnimateDiff~\cite{guo2023animatediff} to inject a motion module to model the temporal dynamics of the 3D flow. We train the motion module layer from scratch, but only insert LoRA (Low-Rank Adaptation) layers~\cite{hu2022lora} into the SD model to ensure that the generative capabilities obtained during pre-training are preserved. In the inference process, the initial points $\mathcal{F}_0$ are obtained through a detector and undergo a corrosion process to prevent interference from points outside the object.

\begin{figure}[t]
  \centering
  \includegraphics[width=0.95\linewidth]{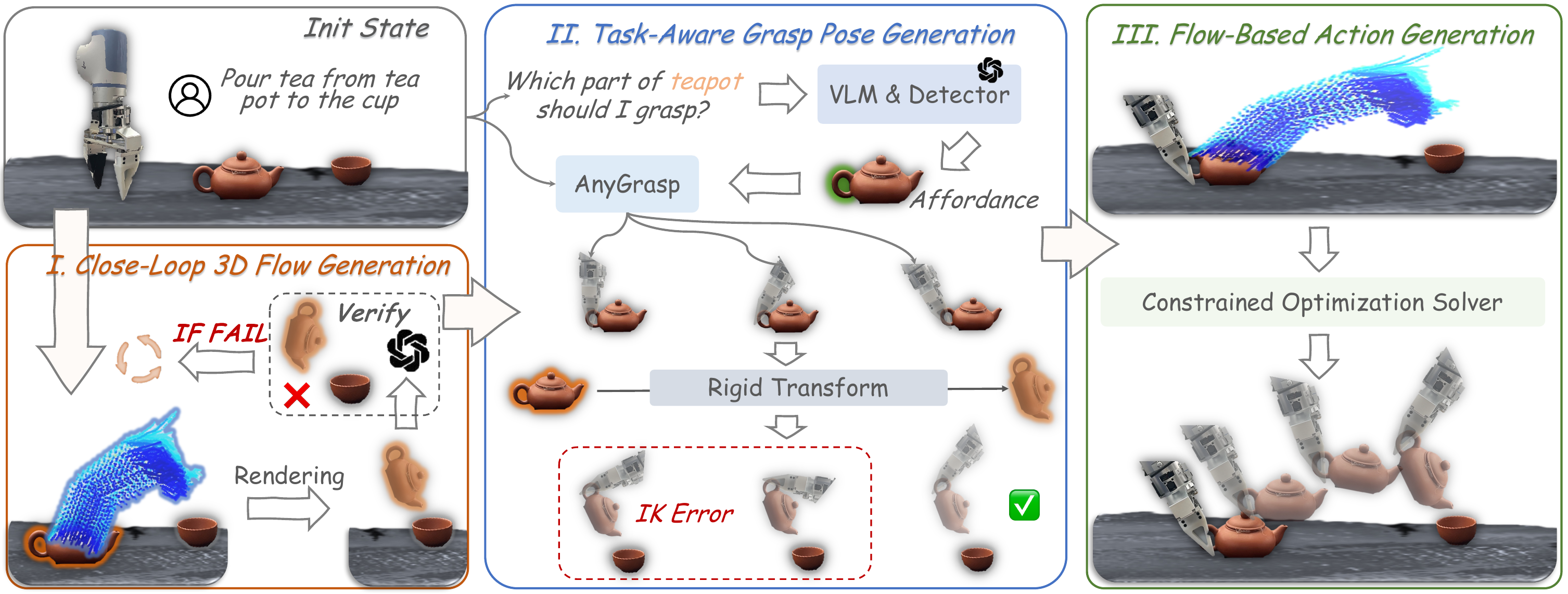}
  \caption{\textbf{Overview of flow-guided action generation pipeline
}. (I) 3DFlowAction first performs closed-loop 3D flow generation through a self-correcting process. (II) A task-aware grasp pose generation process selects a task-relevant grasp pose while avoiding unreachable target positions. (III) An optimization procedure conditioning on 3D flow solves a chunk of actions.}
  \label{fig::flow2act}
\end{figure}

\section{Flow-Guided Action Planning}

Building on the unified action representation from the flow world model, we propose a flow-guided action generation system. This system integrates a target state rendering machine for closed-loop planning, a task-aware grasp pose generation mechanism to select relevant poses and avoid unreachable targets, and a flow condition optimization procedure for cross-embodiment action generation.

\subsection{Closed-Loop Motion Planning for Manipulated Objects}
As the system may sometimes be affected by disturbances, the 3D optical flow predictions of the 3DFlowAction can occasionally be inaccurate. To enhance system stability, we propose an object-centric target state rendering machine as shown in Fig.~\ref{fig::flow2act}(I). Specifically, the optical flow can describe the 3D position of an object at each timestep. We denote the optical flow points from the first timestep as \( P_1 \) and from the last timestep as \( P_2 \). We estimate the transformation matrix \( \mathbf{T} \) between these two sets of points using Singular Value Decomposition (SVD)~\cite{abdi2007singular}:
\[
T = \text{SVD}(P_2, P_1)
\]
We then apply this transformation matrix \( \mathbf{T} \) to the point cloud of the manipulated object at its initial position to obtain the predicted target state of the object's position. The transformed object point cloud is added to the current 3D scene point cloud and reprojected into a 2D image as the predicted output at the moment the task is completed. We input the task instructions and the predicted image into GPT-4o to determine whether a re-prediction of the 3D optical flow is necessary, thereby ensuring the accuracy of the planning process.

\subsection{Task-Aware Grasp Pose Generation}
The selection of grasping poses is also a crucial step for operational tasks. Incorrect grasping poses can sometimes lead to the target position being unreachable or the task being unachievable. We propose a grasping pose generation method based on affordance and predicted 3D flow, as shown in Fig.~\ref{fig::flow2act}(II). Specifically, we will prompt GPT-4o to output the part of the object that should be grasped based on the task instructions, and then use AnyGrasp~\cite{fang2023anygrasp} to generate a series of grasping poses around the object part. Since AnyGrasp is task-unaware, randomly selecting from the candidate grasping poses can sometimes lead to reachability issues for the robotic arm. To address this problem, we apply the transformation matrix \( \mathbf{T} \) extracted from the predicted optical flow to the optimization of the grasping phase. We transform all candidate grasp poses using \( \mathbf{T} \), which indicates the target gripper pose corresponding to the predicted target object position represented by the optical flow. We then utilize the inverse kinematics (IK) of the robotic arm to determine whether these target poses are reachable, thereby selecting the task-related grasp pose.

\subsection{Flow-Based Action Generation}
Since 3D flow captures an object's position in 3D space at each time step, it allows us to represent a manipulation task as a sequence of object poses. This representation enables us to use an optimization procedure to determine the corresponding robot actions, which are expressed as a sequence of end-effector poses in SE(3). The optimization procedure takes the current robot state and constraint function as input, iteratively minimizing the cost of the constraint function to output robot actions that ultimately satisfy the constraint function. Existing methods~\cite{huang2024rekep, huang2024copa, pan2025omnimanip} use code-based constraints generated by GPT-4o, which struggle to describe complex trajectories like rotation in 3D space.
We propose using the 3D optical flow as the constraint function for the optimization procedure, thereby avoiding the cumbersome prompt engineering and inaccurate constraint generation. Specifically, we first select N key points on the surface of the object using farthest point sampling and obtain the corresponding 3D optical flow. Next, we minimize the Euclidean distance in 3D space between the selected initial keypoints and the keypoints corresponding to the predicted flow at the time step t, to obtain the end effector pose for that time step. Finally, we obtain a series of end-effector poses as the final execution actions. More details can be found in Appendix.~\ref{apdix::action_generation}.

\begin{figure}[t]
  \centering
  \includegraphics[width=1\linewidth]{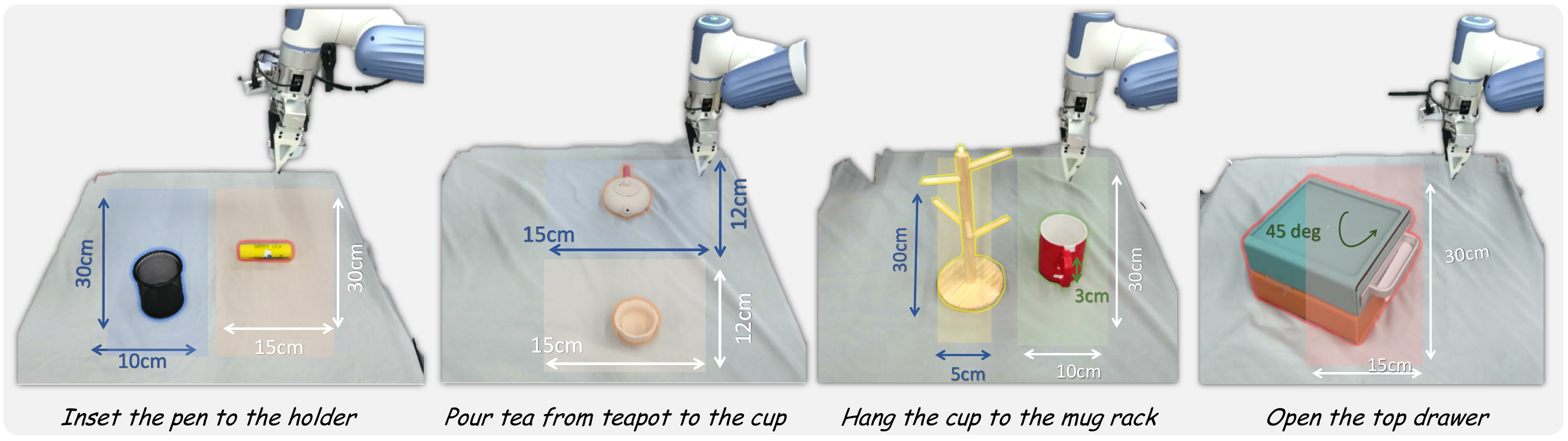}
  \caption{Demonstration of placement of four foundational tasks.}
  \label{fig::exp_setting}
\end{figure}

\section{Experiments} \label{sec::exp}
In this section, we aim to answer the following questions: \textbf{(1)} How does the 3DFlowAction perform compared to existing manipulation world models? \textbf{(2)} Can 3DFlowAction perform cross-embodiment ability? \textbf{(3)} How does the 3DFlowAction compare to current imitation learning methods? \textbf{(4)} How does the 3DFlowAction perform with out-of-domain objects and backgrounds?
 \textbf{(5)} How do the closed-loop planning and large-scale training contribute to its overall performance enhancement?

\subsection{Experimental Setup}
\noindent\textbf{Tasks and Metrics.}
We purposefully select a set of tasks with the goal of examining the capability and generalization of the system we proposed in handling complex tasks, as shown in Fig.~\ref{fig::exp_setting}. 
Pouring tea from the teapot to the cup requires the robot to keep the teapot horizontal and align the spout with the cup's opening. Inserting the pen into the holder involves positioning it vertically and executing complex rotations. Hanging the cup on the mug rack necessitates aligning the handle with the holder. Opening the top drawer must be done without dragging to avoid getting stuck. We also conducted zero-shot testing on four fundamental tasks, each featuring
different target objects and backgrounds, to experimentally evaluate the generalization capabilities. Each setting consists of 10 trials with randomized object poses, and the task success rate is reported. 

\noindent\textbf{Hardware Configuration.}
We conducted experiments on the Xtrainer platform from Dobot. For perception, we employ one Femto Bolt camera, which is positioned opposite the robot to offer a third-person view of the workspace.

\noindent\textbf{Baselines.}
We compare our 3DFlowAction with several baselines, including the baselines of the world model AVDC~\cite{ko2023learning}, Rekep~\cite{huang2024rekep}, as well as imitation learning baselines PI0~\cite{black2024vision} and Im2Flow2Act~\cite{xu2024flow}. 
\textbf{(1)} AVDC~\cite{ko2023learning} employs a video diffusion model to predict future scene changes based on task instructions and scene status, utilizing a 2D optical flow extractor to track moving objects and optimize gripper actions. Similarly. \textbf{(2)} Rekep~\cite{huang2024rekep} uses a Large Vision Language Model (LVLM)~\cite{ren2024grounding} to extract keypoints and generate code-format constraints for target points, followed by an optimization procedure for gripper actions. 
\textbf{(3)} PI0~\cite{black2024vision} adapts a pre-trained VLM for robot control by adding an action expert for continuous actions, fine-tuned on a diverse dataset. \textbf{(4)} Im2Flow2Act~\cite{xu2024flow} learns 2D optical flow generation from a human-collected dataset and trains a diffusion policy using simulated robotic operation data. More details can be found in Appendix.~\ref{apdix::baseline}.

\noindent\textbf{Training data}. We manually collected 30 human demonstrations for each task (without action labels) using human hands, with approximately 10 minutes per task, to fine-tune 3DFlowAction.

\label{ques1}
\subsection{Comparisons with Manipulation World Models}

\begin{table}[t]
\centering
\small
\caption{Comparisons with different world models on four foundational tasks. $^*$ means replacing the learnable action policy with an optimization procedure.}
\label{tab::main}
\resizebox{0.95\columnwidth}{!}{%
\begin{tabular}{lcccc}
\toprule
\textbf{Task} &  {\textbf{AVDC}} &  {\textbf{Rekep}} &  \textbf{Im2Flow2Act$^*$ } &\textbf{3DFlowAction} \\ 
                             \midrule
Pour tea from teapot to the cup &  1/10                               &   2/10                              &                                       2/10&  \textbf{6/10}                                     \\
Insert the pen in holder        &  2/10                              & 1/10                                &                                       2/10&  \textbf{7/10}                                     \\
Hang the cup to the mug rack     & 0/10                               &                                 3/10&                                       0/10&  \textbf{5/10}                                     \\
Open the top drawer                 & 5/10                               & 2/10&                                       6/10&  \textbf{10/10}                                     \\ \midrule
\textbf{Total}                  & 20.0\%                               &                                 20.0\%&                                       25.0\%& \textbf{70.0\%}                                      \\ \bottomrule
\end{tabular}%
}
\end{table}



\begin{table}[t]
\centering
\small
\begin{minipage}{0.38\textwidth}
\caption{3DFlowAction on different robotic arm platforms.}
\label{tab::cross_embodiment}
\resizebox{\textwidth}{!}{%
\begin{tabular}{lcc}
\toprule
\textbf{Task}  & {\textbf{Franka}}& {\textbf{XTrainer}}\\ 
\midrule
Pour tea from teapot to the cup & 7/10  & 6/10 \\
Insert the pen in holder & 7/10 & 7/10 \\
Hang the cup to the mug rack & 4/10 & 5/10 \\
Open the top drawer & 9/10 & 10/10 \\ 
\midrule
\textbf{Total} & 67.5\% & 70.0\% \\ 
\bottomrule
\end{tabular}%
}
\end{minipage}
\hspace{0.02\textwidth} 
\begin{minipage}{0.58\textwidth}
\caption{Comparisons with different imitation learning methods on four foundational tasks.}
\label{tab::imitation_learning}
\resizebox{0.95\textwidth}{!}{%
\begin{tabular}{lccc}
\toprule
\textbf{Task} & {\textbf{PI0}} & {\textbf{Im2Flow2Act}} & {\textbf{3DFlowAction}} \\ 
\midrule
Pour tea from teapot to the cup & 5/10& 4/10  & \textbf{6/10} \\
Insert the pen in holder &   5/10& 2/10 & \textbf{7/10} \\
Hang the cup to the mug rack & 4/10 & 0/10 & \textbf{5/10} \\
Open the top drawer &  6/10& 5/10 & \textbf{10/10} \\ 
\midrule
\textbf{Total} &  50.0\%& 27.5\%  & \textbf{70.0\%} \\ 
\bottomrule
\end{tabular}%
}
\end{minipage}
\end{table}

\noindent\textbf{3DFlowAction demonstrates better effectiveness in capturing the intricate motion of objects in 3D space}. As shown in Tab.~\ref{tab::main}, 
3DFlowAction outperformed the 2D flow world model (Im2Flow2Act$^*$~\cite{xu2024flow}) in all four tasks, highlighting the limitations of 2D optical flow in capturing the complexity of three-dimensional motion trajectories. The reliance on 2D optical flow restricts the optimization process to producing 3D end-effector poses that only reflect object motion within the image plane, often failing to accurately represent true 3D trajectories. These results emphasize the need to transition from a 2D to a 3D framework for better accuracy in object motion capture.

Additionally, 3DFlowAction significantly surpassed the video world model (AVDC~\cite{ko2023learning}) in all tasks. The video model's low resolution and non-object-centric future state generation hinder its effectiveness. High-resolution video generation demands substantial computational resources, while low-resolution outputs can lead to inconsistencies, such as objects disappearing and reappearing unexpectedly, adversely affecting action policies. Moreover, the video model's reliance on external factors complicates future state generation, making it less robust. In contrast, 3DFlowAction generates object-centric 3D motion trajectories, enabling precise descriptions of object motion and posture, thus minimizing the influence of irrelevant factors on downstream action policies.

3DFlowAction also outperformed the VLM code-based world model (Rekep~\cite{huang2024rekep}) across all tasks. The VLM model's reliance on code-based constraints limits its ability to represent complex object motion, as it describes relationships between keypoints primarily in terms of distance. In contrast, 3DFlowAction utilizes optical flow to more effectively and naturally describe the future spatial positions of objects.

\label{ques2}
\subsection{Cross-Embodiment Experiments}
\noindent\textbf{3D optical flow can serve as a unified action feature to bridge different robotic platforms}.
We directly deployed 3DFlowAction on two robotic platforms: Franka~\cite{haddadin2022franka} and XTrainer, without any robot-related fine-tuning. As shown in Tab.~\ref{tab::cross_embodiment}, the performance of 3DFlowAction across different robotic platforms demonstrates its ability to effectively accomplish four complex tasks with consistent performance, all without the need for any new data annotation and model finetuning. This further proves that our method possesses cross-embodiment capabilities, with 3D optical flow serving as an effective unified action representation between different embodiments.

\label{ques3}
\subsection{Comparison with Imitation Learning Methods}

\noindent\textbf{The optimization policy can achieve competitive performance under well-guided action features without the need for teleoperation data}. To ensure a fair comparison, we collected 30 demonstration datasets per task for two imitation learning methods through teleoperation to fine-tune these baselines. As shown in Tab.~\ref{tab::imitation_learning}, 3DFlowAction consistently performs well compared to the two imitation learning methods, thanks to the fact that the input for the optimization policy is 3D optical flow, which provides a comprehensive and effective description of trajectory-related instructions in 3D space.

\begin{table}[t]
\centering
\caption{We evaluate the zero-shot generalization capability of the methods on different target objects and backgrounds.}
\label{tab::generalization}
\resizebox{\columnwidth}{!}{%
\begin{tabular}{lcccccc} 
\toprule
\multirow{3}{*}{\textbf{Task}} & \multicolumn{3}{c}{\textbf{Object Generalization}} & \multicolumn{3}{c}{\textbf{Background Generalization}} \\ 
\cmidrule(lr){2-4} \cmidrule(lr){5-7}
& \textbf{AVDC} & \textbf{PI0} & \textbf{3DFlowAction} & \textbf{AVDC} & \textbf{PI0} & \textbf{3DFlowAction} \\ 
\midrule
Pour tea from teapot to the cup & 0/10 & 3/10 & \textbf{4/10} & 0/10 & \textbf{4/10} & \textbf{4/10} \\ 
Insert the pen in holder & 2/10 & \textbf{6/10} & \textbf{6/10} & 0/10 & 1/10 & \textbf{4/10} \\ 
Hang the cup to the mug rack & 0/10 & 2/10 & \textbf{4/10} & 0/10 & 3/10 & \textbf{4/10} \\ 
Open the top drawer & 4/10 & 5/10 & \textbf{8/10} & 0/10 & 5/10 & \textbf{8/10} \\ 
\midrule
\textbf{Total} & 15.0\% & 40.0\% & \textbf{55.0\%} & 0.0\% & 32.5\% & \textbf{50.0\%} \\ 
\bottomrule
\end{tabular}%
}
\end{table}

\begin{figure}[t]
  \centering
  \includegraphics[width=0.95\linewidth]{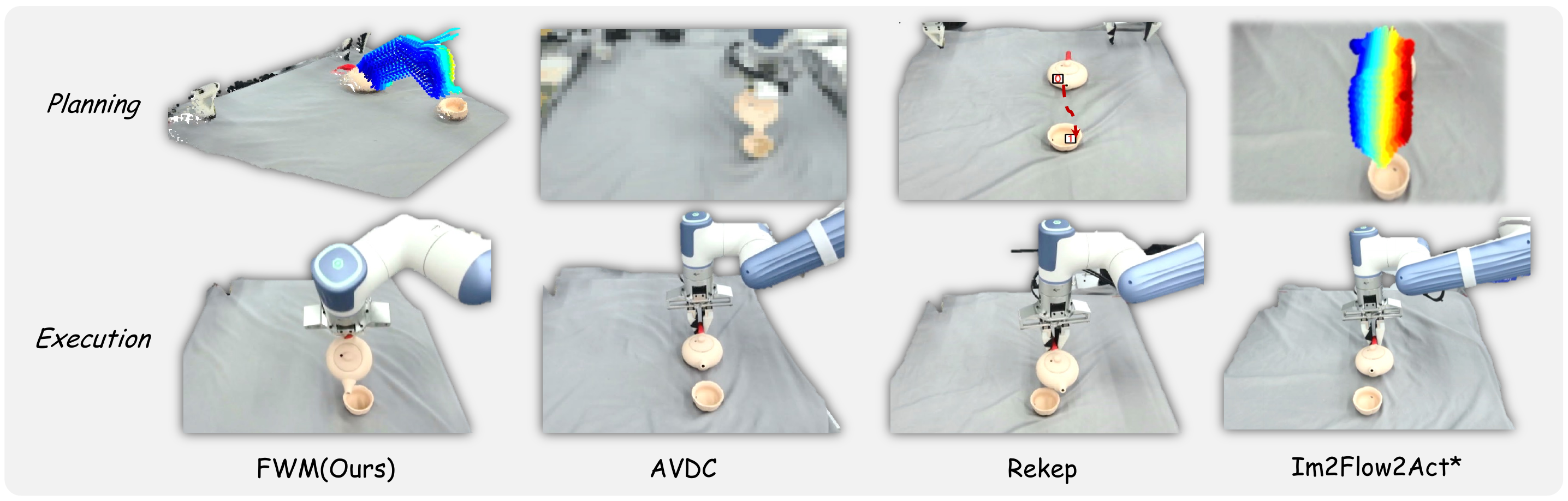}
  \caption{\textbf{Visualization of planning and execution from different world models for pouring tea from the teapot to the cup}. All baseline methods for planning are correct; however, their code-base or 2D planning struggles to fully capture the motion of objects in 3D space, resulting in failures in action planning.}
  \label{fig::l1viz}
\end{figure}

\label{ques4}
\subsection{Experiment on Generalization of Different Objects and Backgrounds}

\noindent\textbf{3DFlowAction can effectively handle the generalization of out-of-domain input due to its object-centric action representation and large-scale pretraining.} As shown in Tab.~\ref{tab::generalization}, the video world model (AVDC) struggles significantly with out-of-distribution data, leading to a drastic decline in success rates compared to in-domain tasks. This limitation arises because AVDC, when simulating future states, must account for background elements that are not relevant to certain tasks. When faced with unseen inputs, the generated video quality deteriorates, making it challenging to effectively guide robotic arm movements. The imitation learning method (PI0) also shows a certain decline in performance when facing out-of-domain scenarios. Our object-centric framework, 3DFlowAction, continues to demonstrate competitive performance across different unseen objects and backgrounds.

\label{ques5}
\subsection{Ablation Studies}
\noindent\textbf{Closed-loop planning.}
We propose a flow-guided rendering mechanism in 3DFlowAction, which renders the final pose of the object to the current state and queries GPT-4o to verify the correctness of the plan. We conduct an ablation study to demonstrate the effectiveness of closed-loop planning. As shown in Tab.~\ref{tab:ablation}, we report the results with closed-loop planning disabled in Variant 1, where the task success rate decreases by an average of 20\% on four manipulation tasks, demonstrating the effectiveness of the closed-loop planning approach.

\noindent\textbf{Large-scale pretraining for 3DFlowAction.}
We propose training the flow world model on the large-scale dataset ManiFlow-110k to learn the physical dynamics of objects as they move in response to different instructions. For a new complex downstream task, we only need to collect 10 to 30 demonstration data points within 10 minutes using human hands, depending on the complexity of the task, without requiring teleoperation of the robotic arm itself. As shown in Tab.~\ref{tab:ablation}, we report the results without large-scale pretraining in Variant 2, where the task success rate decreases by an average of 40\% on four manipulation tasks. It struggles to learn the skills required for downstream tasks and lacks generalization capabilities without pretraining.

\begin{table}[]
\centering
\caption{Impact of closed-loop planning and large-scale pretaining for 3DFlowAction.}
\label{tab:ablation}
\resizebox{0.95\columnwidth}{!}{%
\begin{tabular}{lccccccc}
\toprule
\multirow{2}{*}{\textbf{Methods}} & \textbf{Large-scale }& \textbf{Rendering }& \textbf{Success }&   \textbf{Pour tea from }& \textbf{Insert the pen }& \textbf{Hang the cuo }& \textbf{Open the }\\
 & \textbf{Pretrain}& \textbf{Machine}& \textbf{Rate}& \textbf{teapot to the cup}& \textbf{to the holder}& \textbf{to the mug rack}&\textbf{top drawer}\\ \midrule
\textbf{Variant1}                 &  \with                                              &                                             &       50.0\%  & 3/10 
 & 5/10& 3/10 &                                9/10\\
\textbf{Variant2}                 &                                               & \with                                             &     30.0\%& 3/10& 3/10& 2/10&                                   4/10\\
\textbf{3DFlowAction}                &  \with                                              & \with                                             & \textbf{70.0\%}      & \textbf{6/10}& \textbf{7/10}& \textbf{5/10}&                               \textbf{10/10}\\ \bottomrule
\end{tabular}%
}
\end{table}

\begin{figure}[t]
  \centering
  \includegraphics[width=0.95\linewidth]{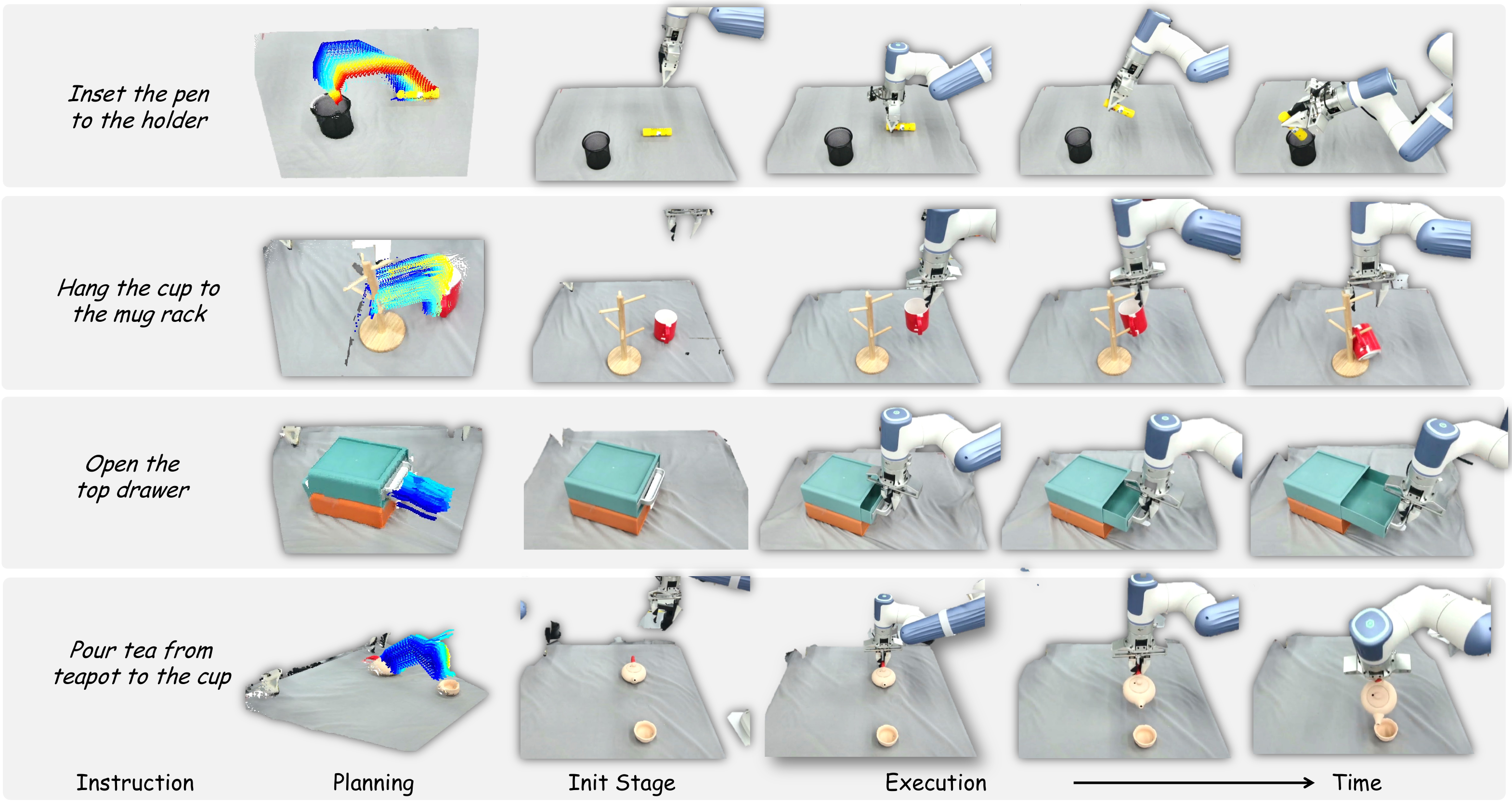}
  \caption{Visualization of 3DFlowAction on four tasks.}
  \label{fig::fwm_viz}
\end{figure}

\section{Conclusion} \label{sec::conclusion}
In this paper, we explore using 3D optical flow as a unified, cross-embodiment, and object-centric action representation to learn the motion patterns of objects based on different instructions from human and robot videos in robot manipulation. To this end, we collected a large-scale 3D flow dataset and trained a flow world model on it. Based on the predicted 3D optical flow, we proposed a rendering machine to assist the model in achieving closed-loop planning to cope with environmental disturbances. The predicted optical flow serves as a constraint for the action policy, outputting a series of execution actions. Experiments demonstrate that our proposed 3DFlowAction exhibits strong generalization across different robotic manipulation tasks and can reliably adapt across embodiments without hardware-specific training.

\noindent\textbf{Limitations.} While advantageous, 3DFlowAction also has limitations. The 3D optical flow faces challenges in modeling the motion of flexible objects due to severe occlusions and complex movements. Additionally, the non-rigid deformations of the objects can lead to the downstream action policy being unable to output effective actions.

\newpage
\newpage

\bibliographystyle{abbrv}
{
	\small
	\bibliography{ref}
}

\newpage
\appendix
\section{Baselines}\label{apdix::baseline}
\noindent\textbf{AVDC~\cite{ko2023learning}} utilizes a video diffusion model to predict future scene state changes based on the current task instructions and scene status. It employs an off-the-shelf 2D optical flow extractor to track moving objects in the predicted video. Finally, based on the 2D keypoint optical flow, an optimization procedure is used to compute a series of gripper end-effector actions. 

\noindent\textbf{Rekep~\cite{huang2024rekep}} first employs a Large Vision Language Model (LVLM)~\cite{ren2024grounding} to extract keypoints from the scene. It then utilizes GPT-4o to generate code-format constraints for multi-stage target points and paths based on different tasks. Finally, using the keypoints and these constraints, an optimization procedure is applied to compute a series of gripper end-effector actions.

\noindent\textbf{Im2Flow2Act$^*$} first predicts the 2D optical flow of object motion based on task instructions using a video diffusion model. Then, based on the 2D keypoint optical flow, an optimization procedure is used to compute a series of gripper end-effector actions. This baseline is modified from Im2Flow2Act~\cite{xu2024flow} by replacing the action policy from diffusion policy to an optimization procedure.

\noindent\textbf{PI0~\cite{black2024vision}} uses a pre-trained VLM as the backbone, which is adapted for robot control by adding a separate action expert that produces continuous actions. This model is pre-trained on a diverse cross-embodiment dataset and can be fine-tuned to adapt to specific operational tasks of the robot.

\noindent\textbf{Im2Flow2Act~\cite{xu2024flow}} first learns the 2D optical flow generation for downstream tasks from a human-collected dataset, and then uses simulated robotic operation data to train the diffusion policy for flow conditions.

\section{Flow-Based Action Generation} \label{apdix::action_generation}
Since 3D flow describes the position of an object in 3D space at each time step, it demonstrates that by representing a manipulation task as a sequence of object poses, we can employ an optimization procedure to solve for robot actions (represented by a sequence of end-effector poses in SE(3)). The optimization procedure takes the current robot state and constraint function as input, iteratively minimizing the cost of the constraint function to output robot actions that ultimately satisfy the constraint function. Existing method~\cite{huang2024rekep} uses constraints generated by GPT based on keypoints; however, experiments~\cite{pan2025omnimanip} have shown that the constraints generated by GPT-4o can only handle relatively simple distance-based tasks and struggle with scenarios involving poor keypoint quality and complex trajectories like rotation.

We propose using the 3D optical flow of object motion as the constraint function for the action solver, thereby avoiding the cumbersome prompt engineering and inaccurate constraint generation required by previous methods. 
We first obtain N keypoints from the predicted optical flow of the first frame using farthest point sampling. Each keypoint $k_i \in \mathbb{R}^3$ refers to a 3D point on the object surface with Cartesian coordinates. An individual solving process can be described as a set of functions $f: \mathbb{R}^{K \times 3} \to \mathbb{R}$ that takes an array of keypoints, represented as $k$, and produces an unbounded cost. When the output satisfies the condition $f(k) \leq 0$, it indicates that the specified constraint is met. This function $f$ is designed as a stateless Python function that utilizes NumPy~\cite{harris2020array} operations to compute the Euclidean distance among the keypoints. In essence, each action-solving process encapsulates several specific spatial relationships between keypoints, which may originate from different components of an object.

An individual solving process adopts a relatively straightforward central idea: it aims for the object to reach the predicted position described by the optical flow at certain time steps. We achieve this goal by minimizing the Euclidean distance in 3D space between the selected initial keypoints and the keypoints corresponding to the predicted optical flow at the time step $t$. Let $k_{\text{initial}}$ represent the selected initial keypoints and $k_{\text{pred}}(t)$ represent the keypoints that come from the predicted optical flow of the Flow World Model at time step $t$. The objective at time step $t$ can be expressed as:
\[
f^{(t)}(k_{\text{initial}}) = \min \sum_{i=1}^{N} \| k_{\text{initial}}^i - k_{\text{pred}}^i(t) \|_2^2
\]
where $N$ is the number of keypoints and $\| \cdot \|_2$ denotes the Euclidean norm. In real-world scenarios, we will also incorporate the robotic arm's inverse kinematics (IK) and collision detection into the objective. Since the object being manipulated and the gripper are rigidly connected, it implies that both share the same trajectory during the place stage. Therefore, we only need to optimize the transformation matrix that represents the position of the object at future time steps relative to its current position. This transformation matrix can be regarded as the motion transformation matrix of the gripper. Ultimately, the gripper trajectory during the place stage is derived from multiple transformation matrices corresponding to future time steps.

\begin{figure}[t]
  \centering
  \includegraphics[width=1\linewidth]{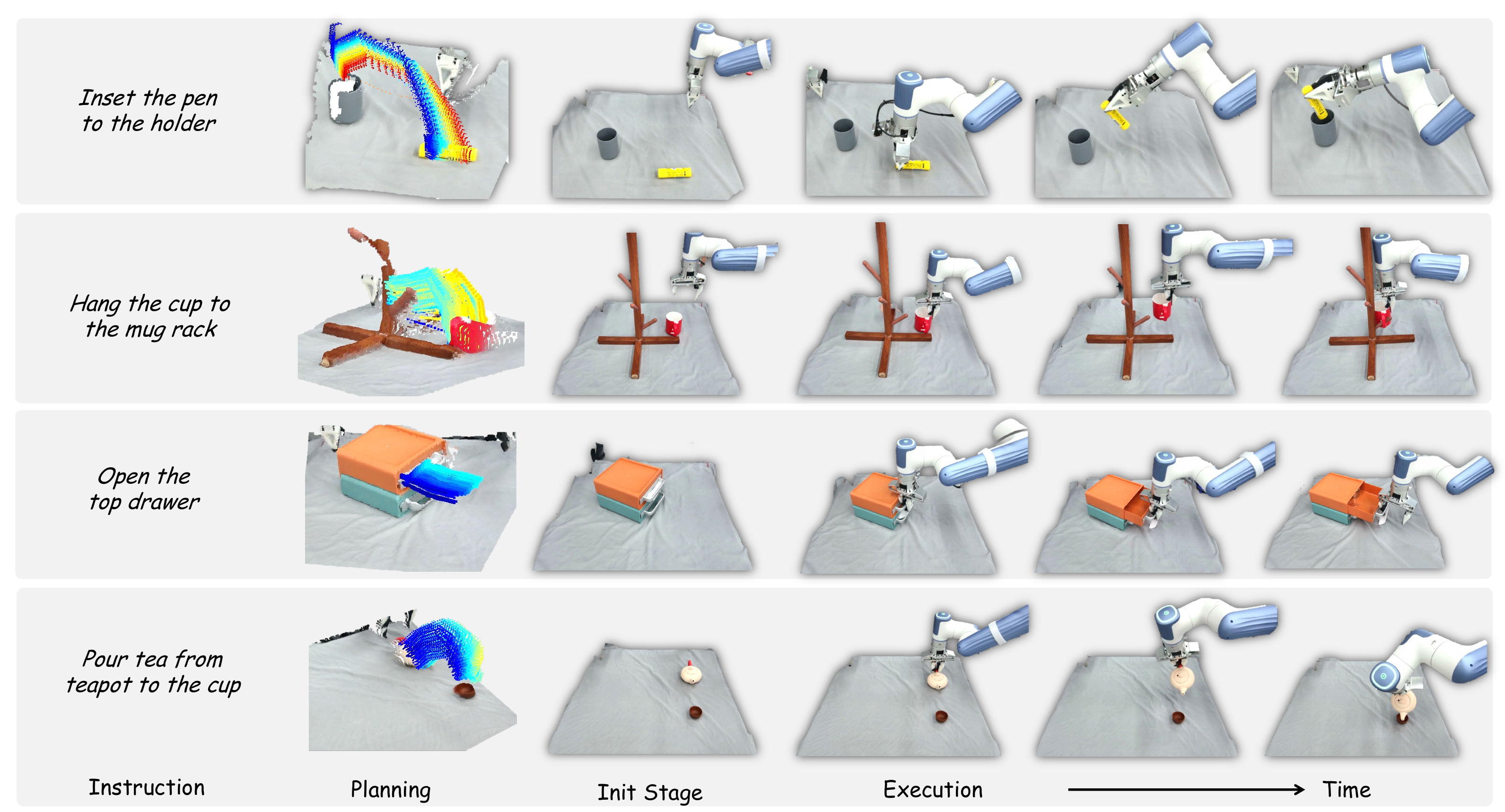}
  \caption{Object generalization experiment visualization.}
  \label{fig::l2viz}
\end{figure}

\begin{figure}[t]
  \centering
  \includegraphics[width=1\linewidth]{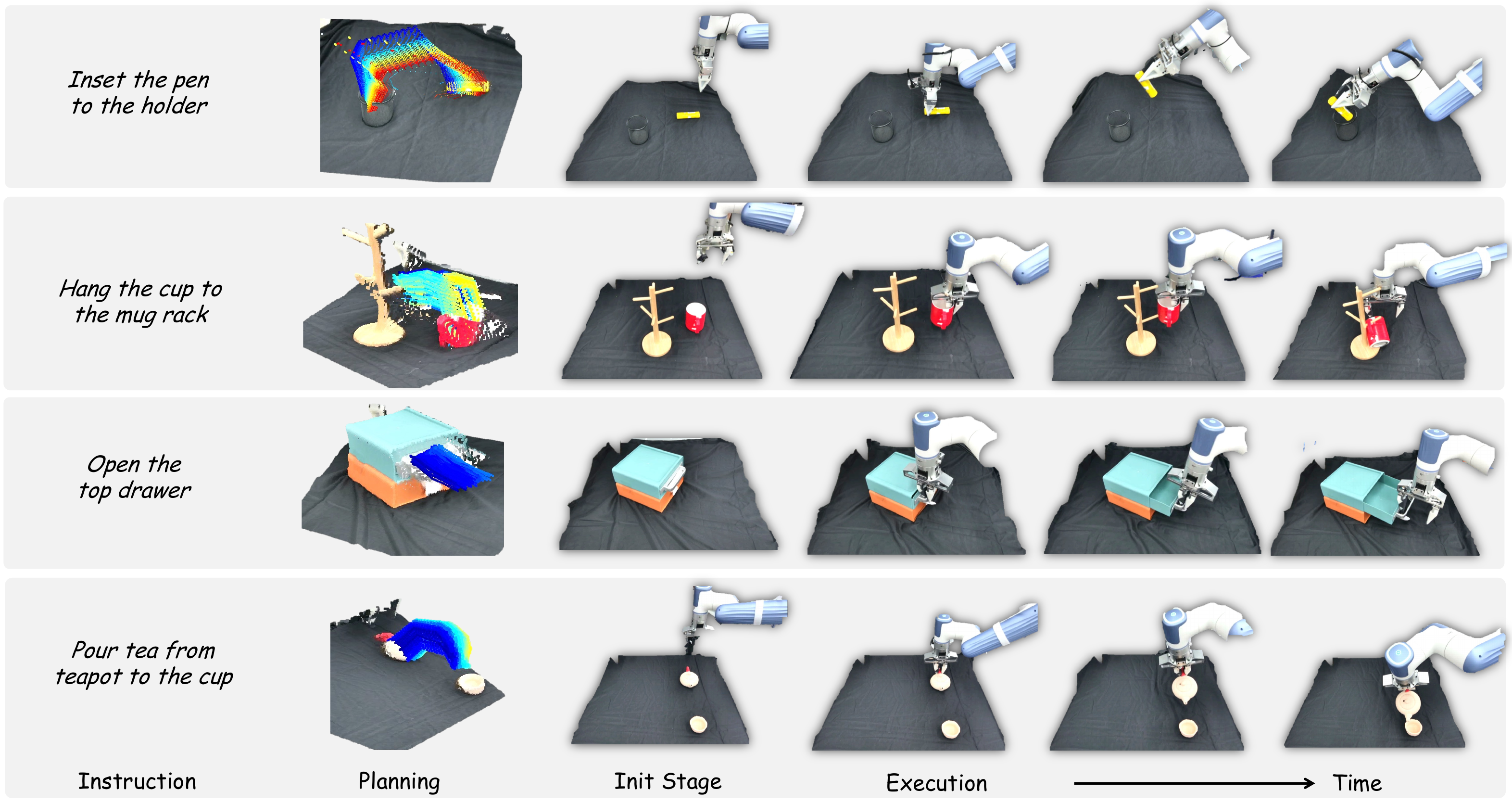}
  \caption{Background generalization experiment visualization.}
  \label{fig::l3viz}
\end{figure}

\begin{figure}[t]
  \centering
  \includegraphics[width=1\linewidth]{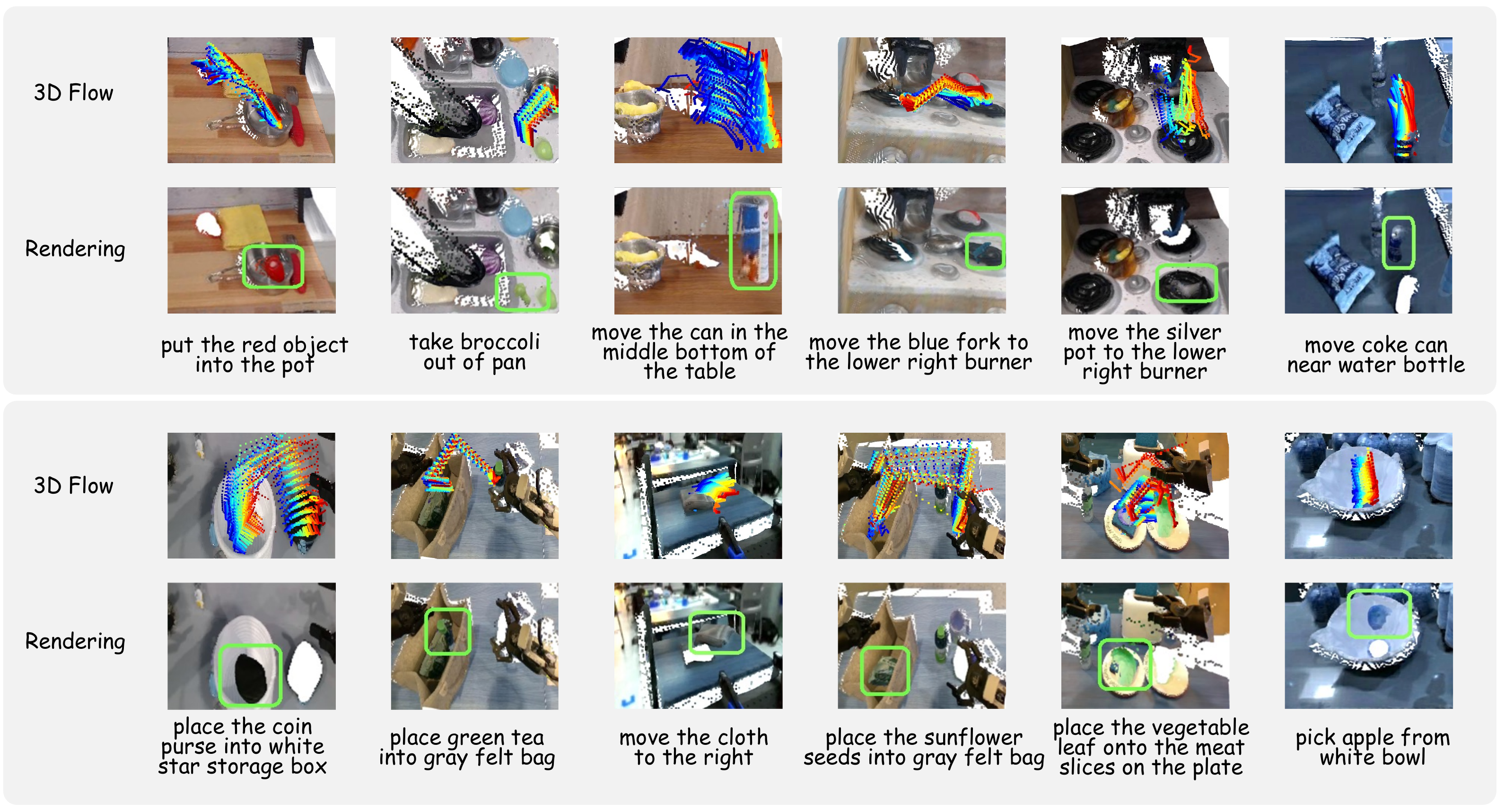}
  \caption{Visualization of in-domain flow generation and target state rendering in ManiFlow-110k.}
  \label{fig::render}
\end{figure}

\section{Implementation Details of Optimization Procedure}
The optimization procedure is addressed and solved using SciPy~\cite{virtanen2020scipy} following Rekep~\cite{huang2024rekep}. For single-arm robots, the decision variable consists of a single end-effector pose, represented by its position and Euler angles $\mathbb{R}^6$. The position terms are constrained by predefined workspace limits, while the rotation terms are restricted to the lower half of the hemisphere, as the end-effector typically faces downward. The decision variables are normalized to the range $[-1, 1]$ based on these bounds.
In the first iteration of solving, the initial guess is set to the current end-effector pose. We employ a sampling-based global optimization method, Dual Annealing \cite{xiang1997generalized}, during this initial iteration to efficiently explore the entire solution space. This is followed by a gradient-based local optimizer, SLSQP \cite{kraft1988software}, which refines the solution. The entire process for this iteration takes approximately one second. In subsequent iterations, we utilize the solution from the previous stage and rely solely on the local optimizer, as it can quickly adapt to minor changes. The optimization is constrained by a fixed time budget, represented as the number of objective function evaluations, to ensure the system operates at a high frequency.

\section{Training Details of 3DFlowAction} \label{apdix::training_details}
We provided the hyperparameters for training the flow world model, as shown in Tab.~\ref{tab::hyparam}, and our pre-trained model was trained on an 8x8 V100 for approximately 2 days.

\begin{table}[ht]
    \centering
    \caption{Model Training Parameters}
    \label{tab::hyparam}
    \begin{tabular}{lc}
        \toprule
        \textbf{Parameter} & \textbf{Value} \\ \midrule
        Learning Rate      & 0.0001          \\
        Batch Size         & 512             \\
        Epochs             & 500            \\
        Optimizer          & AdamW                   \\
        Weight Decay       & 0.01 \\
        Epsilon    & 1e-8 \\
        \bottomrule
    \end{tabular}
\end{table}

\section{More Manipultion Visualization}
We provide more visual results that demonstrate the effective generalization capability of 3DFlowAction across different scenes and tasks involving various target objects, as shown in Fig.~\ref{fig::l2viz}~\ref{fig::l3viz}. The experimental results indicate that 3DFlowAction demonstrates good generalization for both out-of-domain target objects and backgrounds, further highlighting the characteristics of 3D Flow cross-embodiment.

\section{Flow World Model Visualization}
We provided the in-domain prediction results of the flow world model on ManiFlow-110k, and we visualized the predicted optical flow and the corresponding target state rendering results in 3D space, as shown in Fig.~\ref{fig::render}. The experimental results show that 3DFlowAction can learn the motion patterns of objects in 3D space based on different instructions during large-scale pretraining, further demonstrating the effectiveness of large-scale pretraining.

\section{Experimental Tasks}
\noindent\textbf{Pour tea from the teapot to the cup}. This task requires the robot to maintain the teapot in a horizontal position while moving it to avoid spillage. Additionally, the robot must precisely align the spout of the teapot with the approximately 5 cm wide opening of the cup before tilting the teapot to pour the water. 

\noindent\textbf{Insert the pen into the holder}. This task requires the robotic arm to first position the pen in a vertical orientation to create sufficient space for placing it into the holder. The robotic arm must handle complex rotational movements to accomplish this. 

\noindent\textbf{Hang the cup to the mug rack}. This task requires the robotic arm to accurately manage the relative positioning between the cup’s handle and the cup holder.

\noindent\textbf{Open the top drawer}. This task requires the robotic arm to move in accordance with the orientation of the drawer to avoid dragging and getting stuck.


\end{document}